%% file: nn_ann.tex
\documentclass{article} 
\usepackage{iclr2023_conference,times}

\input{math_commands.tex}

\usepackage{hyperref}
\usepackage{url}
\usepackage{graphicx}
\usepackage{booktabs}
\usepackage{tabularx}
\usepackage{amssymb}
\usepackage{here}
\usepackage{dcolumn}

\title{ On Storage Neural Network Augmented \\ Approximate Nearest Neighbor Search}

\iclrfinalcopy
\author{\centerline{Taiga Ikeda \qquad Daisuke Miyashita \qquad Jun Deguchi} \\ \\
\centerline{Kioxia Corporation} \\
}

\begin{document}

\maketitle

\begin{abstract}
Large-scale approximate nearest neighbor search (ANN) has been gaining attention along with the latest machine learning researches employing ANNs. If the data is too large to fit in memory, it is necessary to search for the most similar vectors to a given query vector from the data stored in storage devices, not from that in memory. The storage device such as NAND flash memory has larger capacity than the memory device such as DRAM, but they also have larger latency to read data. Therefore, ANN methods for storage require completely different approaches from conventional in-memory ANN methods. Since the approximation that the time required for search is determined only by the amount of data fetched from storage holds under reasonable assumptions, our goal is to minimize it while maximizing recall. For partitioning-based ANNs, vectors are partitioned into clusters in the index building phase. In the search phase, some of the clusters are chosen, the vectors in the chosen clusters are fetched from storage, and the nearest vector is retrieved from the fetched vectors. Thus, the key point is to accurately select the clusters containing the ground truth nearest neighbor vectors. We accomplish this by proposing a method to predict the correct clusters by means of a neural network that is gradually refined by alternating supervised learning and duplicated cluster assignment. Compared to state-of-the-art SPANN and an exhaustive method using \textit{k}-means clustering and linear search, the proposed method achieves $90\%$ recall on SIFT1M with $80\%$ and $58\%$ less data fetched from storage, respectively.

\end{abstract}

\section{Introduction}
Large-scale Approximate Nearest Neighbor searches (ANNs) for high-dimensional data are receiving growing attentions because of their appearance in emerging directions of deep learning research.
For example, in natural language processing, methods leveraging relevant documents retrieval by similar dense vector search have significantly improved the scores of open-domain question-answering tasks~\citep{DPR}.
Also for language modeling tasks, \citet{Retro} showed a model augmented by retrieval from 2 trillion tokens performs as well as 25 times larger models. 
In computer vision, \citet{Nakata} showed that image classification using ANNs has potential to alleviate catastrophic forgetting and improves accuracy in continual learning scenarios.
In reinforcement learning, exploiting past experiences stored in external memory for an agent to make better decisions has been explored~\citep{MFEC, NEC}. Recently, \citet{RARL} and \citet{LargeScaleRetrievalRL} attempted to scale up the capacity of memory with the help of ANN and showed promising results.

ANNs are algorithms to find one or \textit{k} key vectors that are the nearest to a given query vector among a large number of key vectors. Strict search is not required but higher recall with lower latency is demanded. In order to achieve this, an index is generally build by data-dependent preprocessing. Thus, an ANN method consists of the index building phase and the search phase.
As the number of key vectors increases, storing all of them in memory (e.g. DRAM) becomes very expensive, and it is forced to store the vectors in storage devices such as NAND flash memory.
In general, a storage device has much larger capacity per cost, but also its latency is much larger than memory.
When all the key vectors are stored in memory, as seen in the most papers regarding ANN, it is effective to reduce the number of calculating distance between vectors by employing graph-based~\citep{HNSW,NSG} or partitioning-based methods~\citep{NeuralLSH} and/or to reduce the time for each distance calculation by employing quantization techniques such as Product Quantization~\citep{PQ}.
On the other hand, when the key vectors are stored in storage rather than memory, the latency for fetching data from the storage becomes the dominant contributor in the total search latency.
Therefore, the ANN method for the latter case (ANN method for storage) needs a completely different approach from the ANN method for the former case (all-in-memory ANN method). In this paper, we identify the most fundamental challenges of the ANN method for storage, and explore ways to solve them.

Since the latency for fetching data is approximately proportional to the amount of fetched data, it should be good strategy to reduce the number of vectors to be fetched during search as much as possible, while achieving high recall.
Although SPANN~\citep{SPANN}, which is the state-of-the-art ANN method for storage, is also designed with the same strategy, our investigation reveals that the characteristics of its index are still suboptimal from the viewpoint of the number of fetched vectors under a given recall.
Moreover, perhaps surprisingly, an exhaustive method combining simple \textit{k}-means clustering and linear search can perform better than SPANN on some dataset in terms of this metrics.

Another consideration worth noting is the exploitation of the efficient and high-throughput computation of GPUs. It is reasonable to assume that an ANN algorithm used in deep learning application runs on the same system as the deep learning algorithm runs on, and most deep learning algorithms are designed to be run on the system equipped with GPUs. Therefore, the ANN algorithm will be also run on the system with GPU.

In partitioning-based ANN methods, key vectors are partitioned into clusters, which are referred to as posting lists in SPANN paper, usually based on their proximity in the index building phase. First, in the search phase, some clusters that would contain the desired vector with high probability are chosen according to the distance between the query vector and the representative vector of each cluster. Second, the vectors in the chosen clusters are fetched from storage. Since the page size of storage devices is relatively large (e.g. 4KB), it is efficient to make a page contain vectors in the same cluster, as other ANN methods for storage~\citep{SPANN, DiskANN} also adopt. Third, by computing distances between the query and the fetched vectors, \textit{k} closest vectors are identified and output.
In order to achieve high recall with low latency, we need to increase the accuracy to choose the correct cluster containing the ground truth nearest vector at the first step of the search phase.

If clustering is made by \textit{k}-means algorithm and a query vector is picked from the existing key vectors, the cluster whose centroid vector is the closest to the query always contains the nearest neighbor key vector by the definition of \textit{k}-means algorithm.
However, when a query is not exactly same as any one of key vectors, which is common case for ANN, often the cluster whose representative vector is the closest to the query does not contain the nearest neighbor vector, and this limits recall.
To the best of our knowledge, this problem has not been explicitly addressed in the literatures.
Our intuition to tackle with this problem is that the border lines that determine which cluster contains the nearest neighbor vector of a query are different from and more complicated than the border lines that determine the assignment of clusters to key vectors. We employ neural networks trained on given clustered key vectors to predict the correct cluster among the clusters defined by the complicated border lines.
Assigning multiple clusters to a key vector is a conventional method to improve the accuracy to choose the correct cluster. Combining this duplication technique with our method could provide the additional effect of relaxing demands on the neural network by simplifying border lines.
We demonstrate how our method works with visualization using 2-D toy data, and then empirically show that it is effective for realistic data as well.
Our contributions include:
\begin{itemize}
   \item We clarify that we need to reduce the amount of data fetched from storage when key vectors are sit in storage devices, because the latency for fetching data is dominant part of the search latency. 
   \item We first explicitly point out and address the problem that, in partitioning-based ANN method, often the nearest cluster does not contain the nearest neighbor vector of a query vector and this limits the recall-latency performance.
   \item We propose a new ANN method combining cluster prediction with neural networks and duplicated cluster assignment, and show empirically that the proposed method improves the performance on two realistic million-scale datasets. 
\end{itemize}

\section{Related works}
\textbf{ANN for storage.} 
DiskANN~\citep{DiskANN} is a graph-based ANN method for storage. 
The information of connections defining graph structure and the full precision vectors are stored on storage and the vectors compressed by Product Quantization~\citep{PQ} are stored in memory.
The algorithm traverses the graph by reading the connection information only on the path from storage and computes distances between a query and the compressed vectors in memory.
Although they compensate the deterioration of recall due to lossy compression by combining reranking using full precision vector data, the recall-latency performance is inferior to SPANN~\citep{SPANN}.
SPANN is another method dedicated to ANN for storage and exhibits state-of-the-art performance. It employs a partitioning-based approach.
By increasing the number of clusters as much as possible, it achieved to reduce the number of vectors fetched from storage under a given recall. In order to reduce the latency to choose clusters during search even when the number of clusters are large, they employ SPTAG algorithm that combines tree-based and graph-based ANNs.
They also proposed an efficient duplication method aiming at increasing probability that a chosen cluster contains the ground truth key vector.
However, our investigation in Section~\ref{metrics_memory} shows that its performance can be worse than a naive exhaustive method on some dataset.

\textbf{ANN with GPU.}
FAISS~\citep{FAISS} supports a lot of ANN algorithms accelerated by using GPU's massively parallel computing. On-storage search is also discussed in their project page.
SONG~\citep{SONG} optimized the graph-based ANN algorithm for GPU. They modified the algorithm so that distance computations can be parallelized as much as possible, and showed significant speedup. However, they assume only all-in-memory scenarios.

\textbf{ANN with neural networks.}
DSI~\citep{DSI} predicts the indices of the nearest neighbor key vectors directly from the query vectors with a neural network. We explore to use neural networks to predict the clusters containing the nearest neighbor vector rather than the vector indices themselves. DSI is also targeted for all-in-memory ANN.
BLISS~\citep{BLISS} and NeuralLSH~\citep{NeuralLSH} are methods to improve the partitioning rule using neural networks. They apply the same rule to a query for choosing clusters as well. As depicted in Section~\ref{visualization}, when the rule for partitioning keys is employed to choose clusters, often the chosen clusters don't contain the ground truth key vector. Our method where a neural network is trained to predict the correct cluster for a given query is orthogonal and can be combined with these methods.

\section{Preliminaries}
\subsection{System environment}
In this paper, we assume that the system on which our ANN algorithm runs has GPUs and storage devices in addition to CPUs and memories. The GPU provides high-throughput computing through massively parallel processing. The storage can store larger amount of data at lower cost, but has larger read latency than memory devices. Data that are commonly used for all queries, e.g., all the representative vectors of clusters, are loaded in advance on the memories from which CPU or GPU can read data with low latency and is always there during the search. On the other hand, all the key vectors are stored in the storage devices, and for simplicity, we assume that data fetched from storage for computations for a query is not cached on memory to be reused for computations for another query.

\subsection{Metrics}
\label{metrics}

\subsubsection{The number of fetched vectors as a proxy metrics of latency}

Our goal is to minimize the average search time per query for nearest neighbor search, which we refer to as mean latency, and simultaneously to maximize the recall.
Without loss of generality, the mean latency $T$ in the systems described in the previous subsection is expressed by the following equation,
$$ T = T_a + T_b + T_c, $$
where $T_a$ is the latency for computations using data that are always sit in memory, $T_b$ is the latency required for fetching data from storage, and $T_c$ is the latency for computations using data fetched from storage for each query.
For example, in a partitioning-based ANN method such as SPANN, $T_a$ is the latency for the process to determine the clusters (called as the posting lists in SPANN) to be fetched from storage, $T_b$ is the latency for fetching the vectors in the chosen clusters from storage, and $T_c$ is the latency for the computations to find the nearest neighbor vectors in the fetched vectors.
In this paper, for simplicity, assuming that $T_a \ll T_b$ and $T_c \ll T_b$, we employ the following approximation,
$$ T \approx T_b. $$
Then, since $T_b$ is roughly proportional to the number of fetched vectors, \textit{the number of fetched vectors} is an effective metrics to evaluate the mean latency.

The above assumptions are reasonable in a realistic setting.
For $T_c$, the computing performances of CPUs equipped with vector arithmetic units and GPUs capable of massively parallel operations range from several hundred GFLOPS to more than TFLOPS. On the other hand, read bandwidth of storage devices is at best a few GB/s even when high-speed NVMe is used. This means the fetched data in $T_b$ can be processed in less than $1 / 100$ of $T_b$. Note that if the most of the process for $T_c$ is executed in parallel with the process for $T_b$, for example, by performing distance calculation in the background of asynchronous storage access, the effective $T_c$ becomes almost zero.
For $T_a$, when an exhaustive linear search is used to choose the clusters, i.e., calculating the distance between the query and the representative vectors of all clusters in order to find the closest clusters, a 10-TFLOPS GPU can process 10 million representative vectors of 100 dimension each within a much shorter time than $T_b = 1$ ms.
Also in the SPANN case without GPUs, since a fast algorithm that combines tree-based method and graph-based method is applied to choose the clusters, $T_a$ is quite short even when the number of posting lists is as large as a few hundred million.
As a typical example, Table~\ref{latency} shows the measured $T_a$, $T_b$, and $T_c$ on SIFT1M dataset.

\begin{table}[h]
   \centering
   \begin{minipage}{0.48\linewidth}
      \centering
      \begin{tabularx}{\linewidth}{llll}
         \toprule
          & \multicolumn{3}{c}{\#clusters} \\
          \multicolumn{1}{c}{\textbf{Method}} & \multicolumn{1}{c}{\textbf{1K}} & \multicolumn{1}{c}{\textbf{10K}} & \multicolumn{1}{c}{\textbf{100K}} \\
         \midrule
            GPU & $<$0.0002 & $<$0.0004 & $<$0.004 \\
            SPANN & $<$0.24 & $<$0.27 & $<$0.29 \\
         \bottomrule
      \end{tabularx}
   \end{minipage}
   \hspace{0.01\linewidth}
   \begin{minipage}{0.48\linewidth}
      \centering
      \begin{tabularx}{\linewidth}{llll}
         \toprule
            & \multicolumn{3}{c}{\#vectors fetched from storage} \\
            & \multicolumn{1}{c}{\textbf{1k}} & \multicolumn{1}{c}{\textbf{10K}} & \multicolumn{1}{c}{\textbf{100K}} \\
         \midrule
            $T_b$ & $>$1 & $>$10 & $>$100 \\
            \midrule
            $T_c$, CPU & $<$0.0005 & $<$0.003 & $<$0.03 \\
            $T_c$, GPU & $<$0.0002 & $<$0.0004 & $<$0.004 \\
         \bottomrule
      \end{tabularx}
   \end{minipage}
   \caption{Left: $T_a$ in millisecond. For GPU, $T_a$ is measured using \texttt{FlatIndex} of FAISS. Right: $T_b$ and $T_c$ in millisecond. $T_b$ is measured using SPANN implementation. $T_c$ on CPU and GPU are both measured using \texttt{FlatIndex} of FAISS.}
   \label{latency}
\end{table}

\subsubsection{Memory usage}
\label{metrics_memory}
\begin{figure}[h]
   \begin{minipage}[h]{0.48\linewidth}
      \includegraphics[keepaspectratio, width=\linewidth]{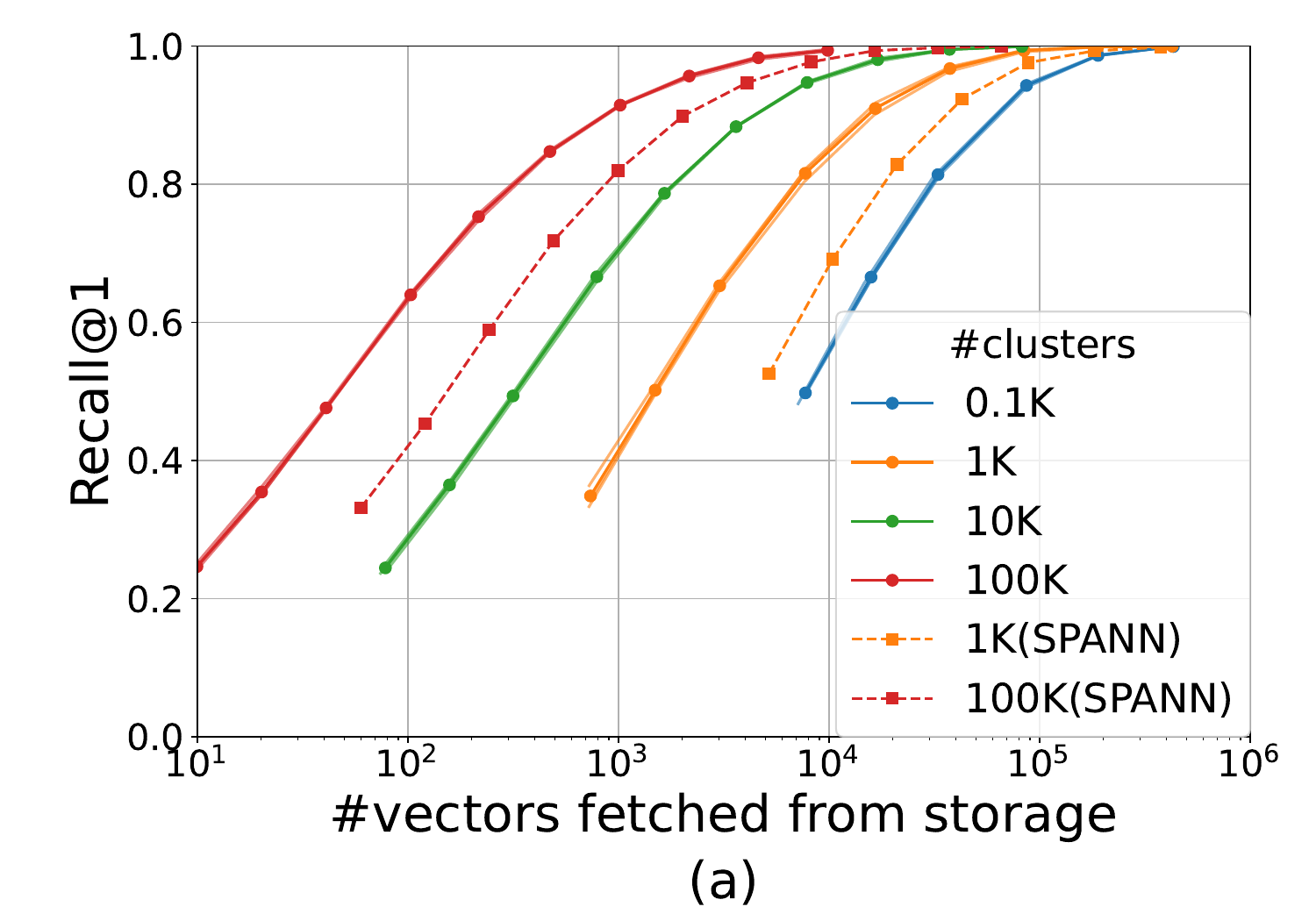}
   \end{minipage}
   \hspace{0.01\linewidth}
   \begin{minipage}[h]{0.48\linewidth}
      \includegraphics[keepaspectratio, width=\linewidth]{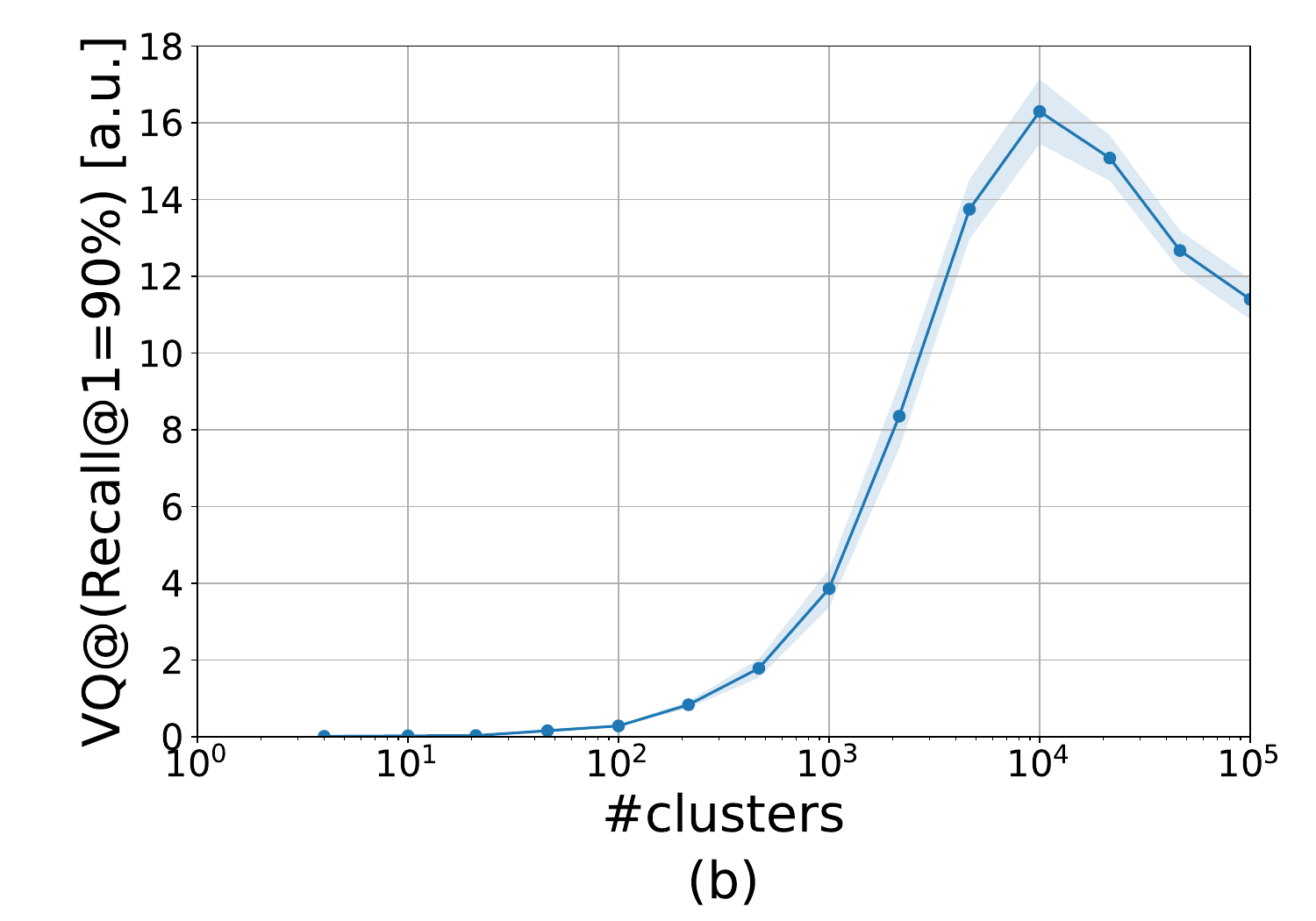}
   \end{minipage}
\caption{(a) Recall@1 vs the number of vectors fetched from storage. It greatly depends on the number of clusters. (b) VQ under recall@1=90\% vs the number of clusters. The line and error bar shows the average and standard deviation of 10 measurements. In these experiments, we use one million SIFT1M base data as key vectors and ten thousand SIFT1B query data as a query vectors.}
\label{fig_metrics}
\end{figure}

Since memory usage greatly affects the latency of in-memory ANNs, VQ (Vector-Query), which is a measure of throughput normalized by the memory usage, is introduced in GRIP~\citep{GRIP} to compare algorithms with different memory usage as fairly as possible, and is also utilized in SPANN~\citep{SPANN}.
SPANN claims superior capacity in large vector search scenarios because this VQ value is greater than that of other algorithms.
Here, we consider whether VQ is really fair metrics for comparing the methods with different memory usage.
In a partitioning-based ANN method for storage, memory capacity limits the number of clusters since the representative vectors of all the clusters must be kept in memory during search.
Then, we investigate how the number of clusters affects the recall-latency and recall-VQ curves.
Figure~\ref{fig_metrics}(a) shows the dependency of recall versus the number of fetched vectors when the simplest \textit{k}-means and linear search method (\texttt{IVFFlatIndex} of FAISS) are utilized and it is clear that the number of fetched vectors significantly decreases as the number of clusters increases.
As shown in Figure~\ref{fig_metrics}(b), even when we use VQ metrics, the VQ value under recall@1=90\% greatly varies depending on the number of clusters. This indicates that VQ is not a suitable metrics for comparison between algorithms with different memory usage.
Based on the above discussion, this paper evaluates ANN methods for storage using \textit{the number of fetched vectors under a given recall and a given number of clusters}.

Another finding by this investigation is that the number of fetched vectors under a given recall of SPANN is not always better than that of the exhaustive method as shown in Figure~\ref{fig_metrics}(a).

\subsection{Visualization with 2-dimensional toy data}
\label{visualization}

\begin{figure}[h]
      \begin{minipage}[h]{0.48\linewidth}
         \includegraphics[keepaspectratio, width=\linewidth]{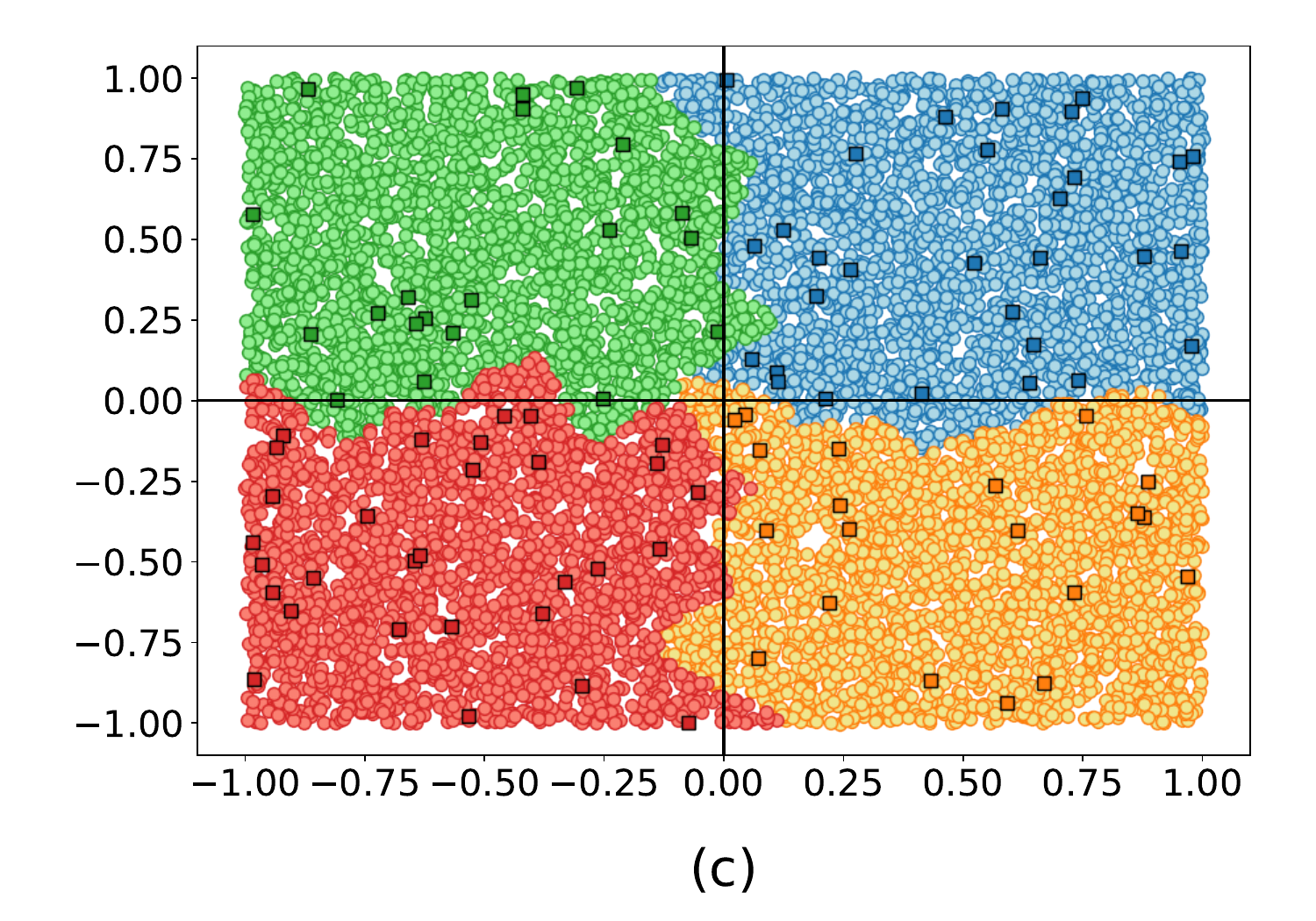}
      \end{minipage}
      \hspace{0.01\linewidth}
      \begin{minipage}[h]{0.48\linewidth}
         \includegraphics[keepaspectratio, width=\linewidth]{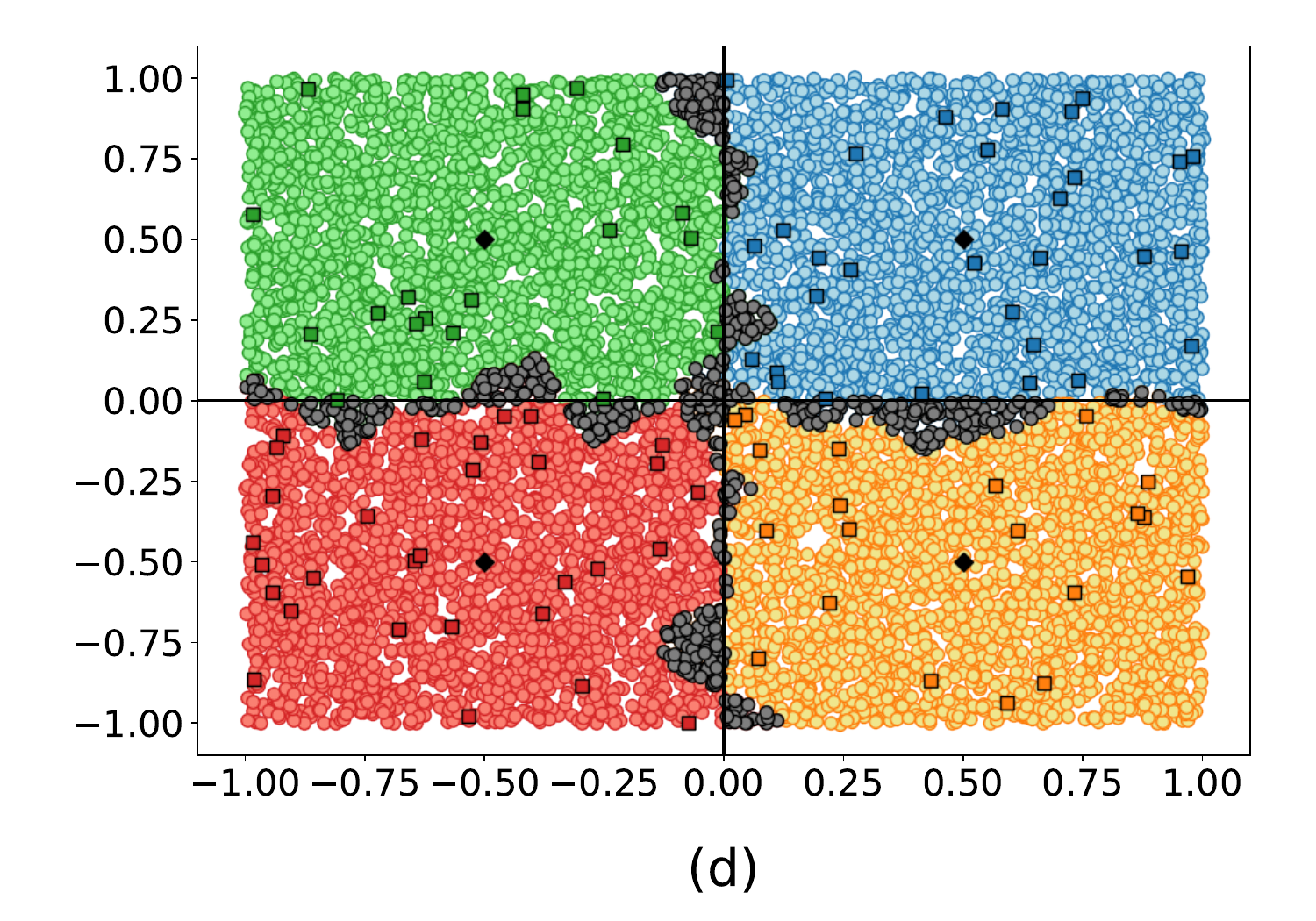}
      \end{minipage}
   \caption{Visualization with 2-dimensional toy data. (a) \textit{Key vectors} are partitioned into four clusters. The cluster assignment is expressed by color. (b) \textit{Query vectors} colored by the \textit{chosen} cluster in the search phase by the conventional method. The query vectors are shown in light-colored circle. (c) \textit{Query vectors} colored by the \textit{correct} cluster that contains the nearest key vector to each query vector. The query vectors are shown in light-colored circle and the key vectors are shown in dark-colored rectangle. (d) Wrong choices are shown in gray.}
   \label{fig_2d_explain}
\end{figure}

In this section, we explain the intuition behind our proposed method and visually demonstrate how it improves the accuracy to choose the correct cluster.
One hundred key vectors are uniformly sampled from 2-dimensional space between $-1$ to $1$.
In the index building phase, we divide the key vectors into four clusters. The representative vectors of the clusters are placed at $(x,y) = (1/2, 1/2), (-1/2, 1/2), (-1/2, -1/2), (1/2, -1/2)$. Then we assign one cluster to each key vector according to the Euclidean distance, i.e., the distance of a key vector to the representative vector of the assigned cluster is smaller than that of other clusters.
Figure~\ref{fig_2d_explain}(a) shows the key vectors colored by the assigned cluster.
In a conventional method, when a query is given in the search phase, we choose one cluster whose representative vector is the closest to the query among the four clusters, which is the same manner as that used for assigning clusters to key vectors in the index building phase. Figure~\ref{fig_2d_explain}(b) shows $10000$ queries colored by the cluster chosen for each query, and the clusters are clearly divided into four quadrants as expected.
On the other hand, Figure~\ref{fig_2d_explain}(c) shows the queries colored by the correct cluster containing the nearest neighbor key vector of each query.
We can see that the nearest neighbor key vector of a query vector in the first quadrant, $x>0,y>0$, can be contained in the cluster in green whose representative vector is in the second quadrant, $x<0,y>0$. As a result, the true border lines of the clusters for query are quite complex.
In Figure~\ref{fig_2d_explain}(d), queries where wrong clusters are chosen are shown in gray. When the chosen cluster does not contain the nearest neighbor vector, we need to fetch vectors from another cluster to increase the recall. This phenomenon leads to the increase in the number of fetched vectors under a given recall, and the deterioration of recall-latency tradeoff.

\begin{figure}[h]
      \begin{minipage}[h]{0.48\linewidth}
         \includegraphics[keepaspectratio, width=\linewidth]{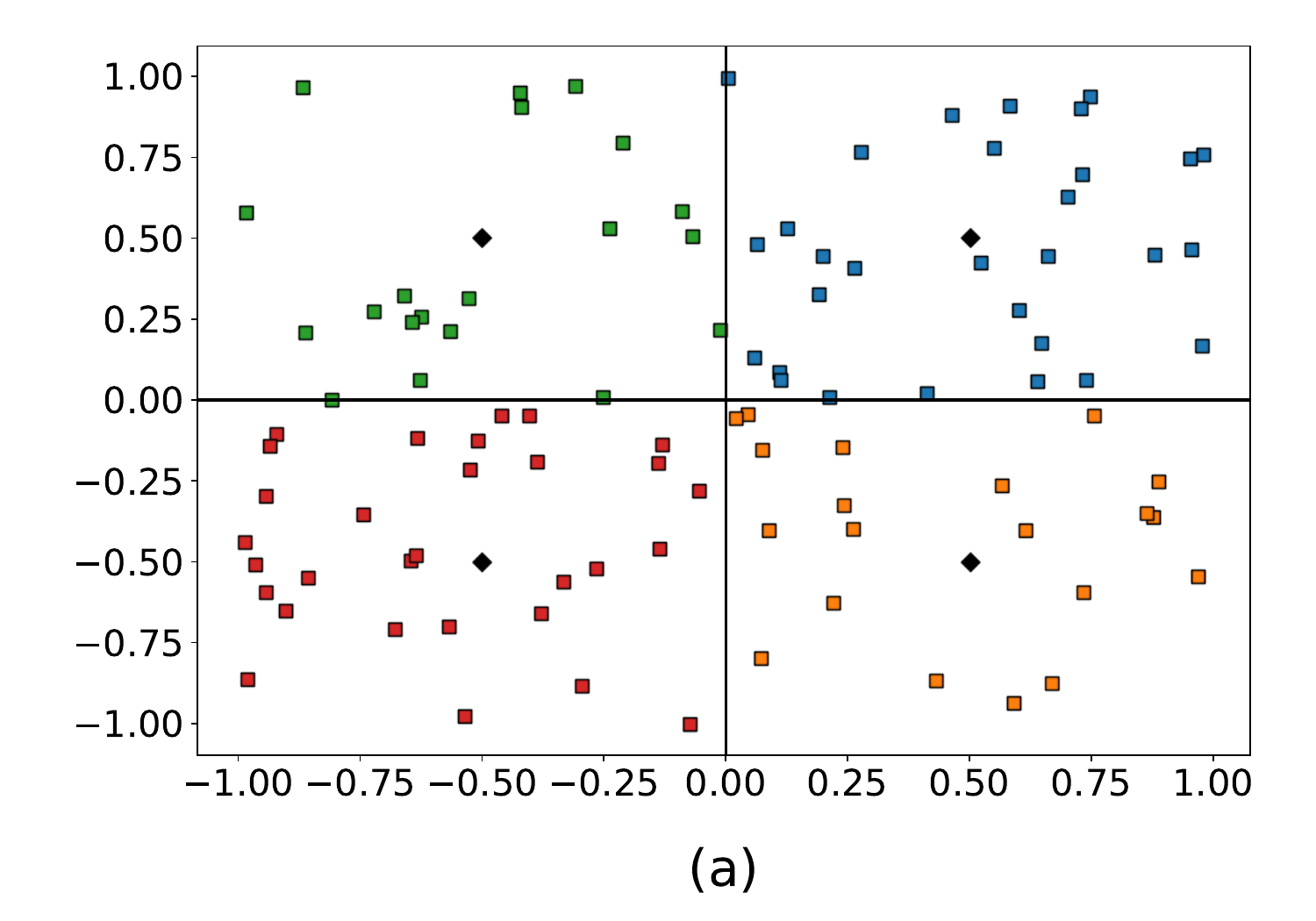}
      \end{minipage}
      \hspace{0.01\linewidth}
      \begin{minipage}[h]{0.48\linewidth}
         \includegraphics[keepaspectratio, width=\linewidth]{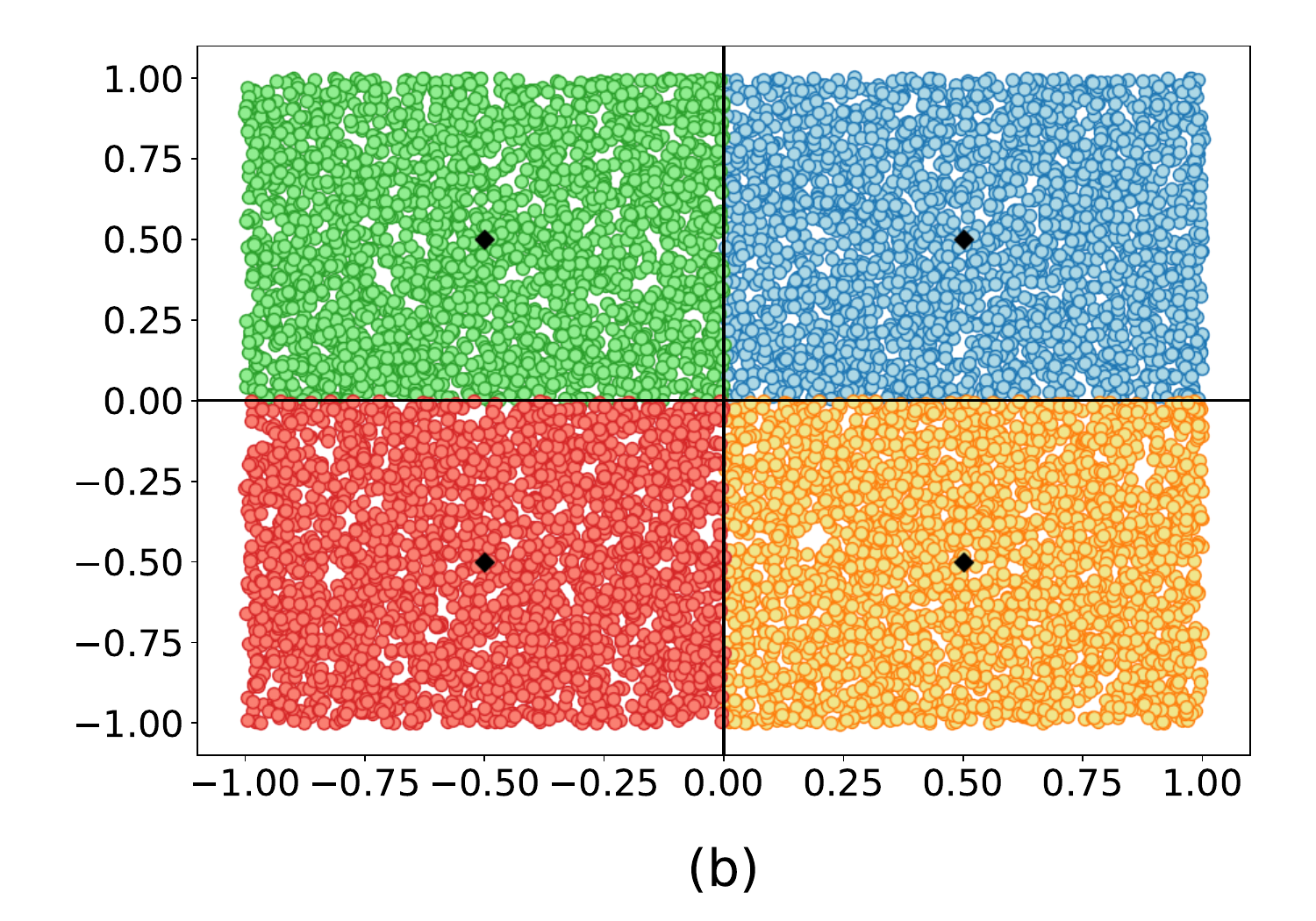}
      \end{minipage}
\end{figure}

\section{Proposed method}

\begin{figure}[h]
      \begin{minipage}[h]{0.32\linewidth}
         \includegraphics[keepaspectratio, width=\linewidth]{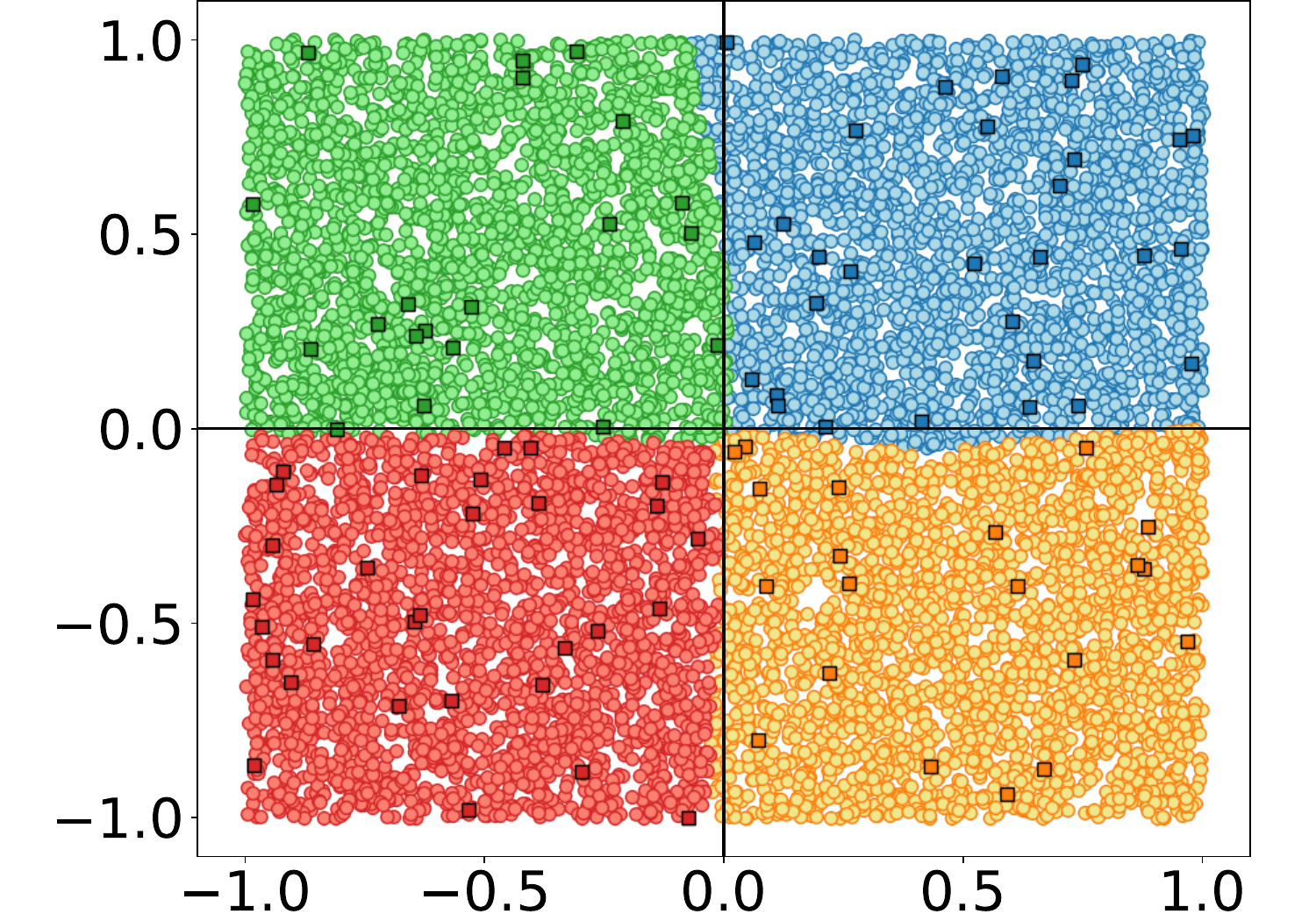}
      \end{minipage}
      \hspace{0.01\linewidth}
      \begin{minipage}[h]{0.32\linewidth}
         \includegraphics[keepaspectratio, width=\linewidth]{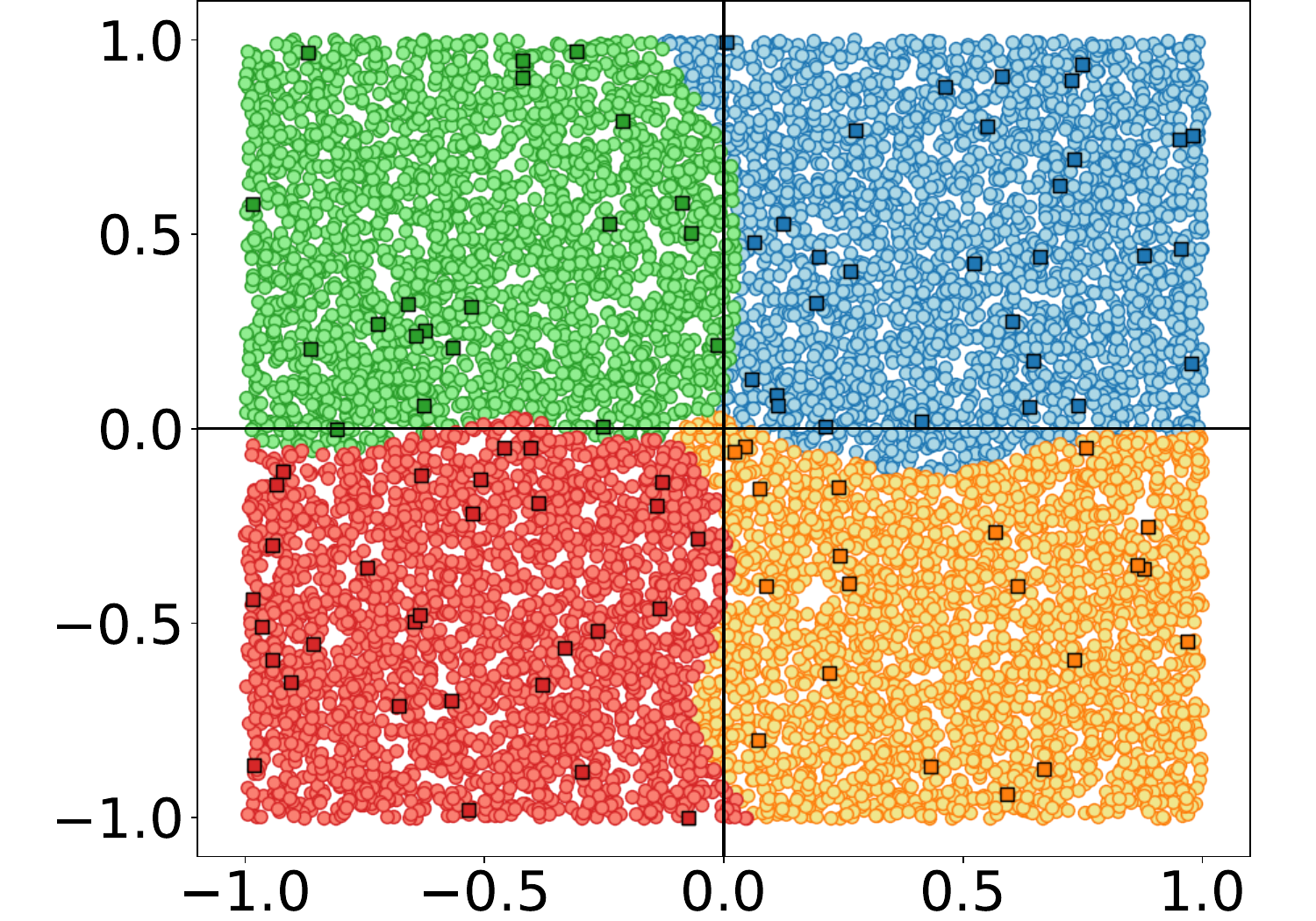}
      \end{minipage}
      \hspace{0.01\linewidth}
      \begin{minipage}[h]{0.32\linewidth}
         \includegraphics[keepaspectratio, width=\linewidth]{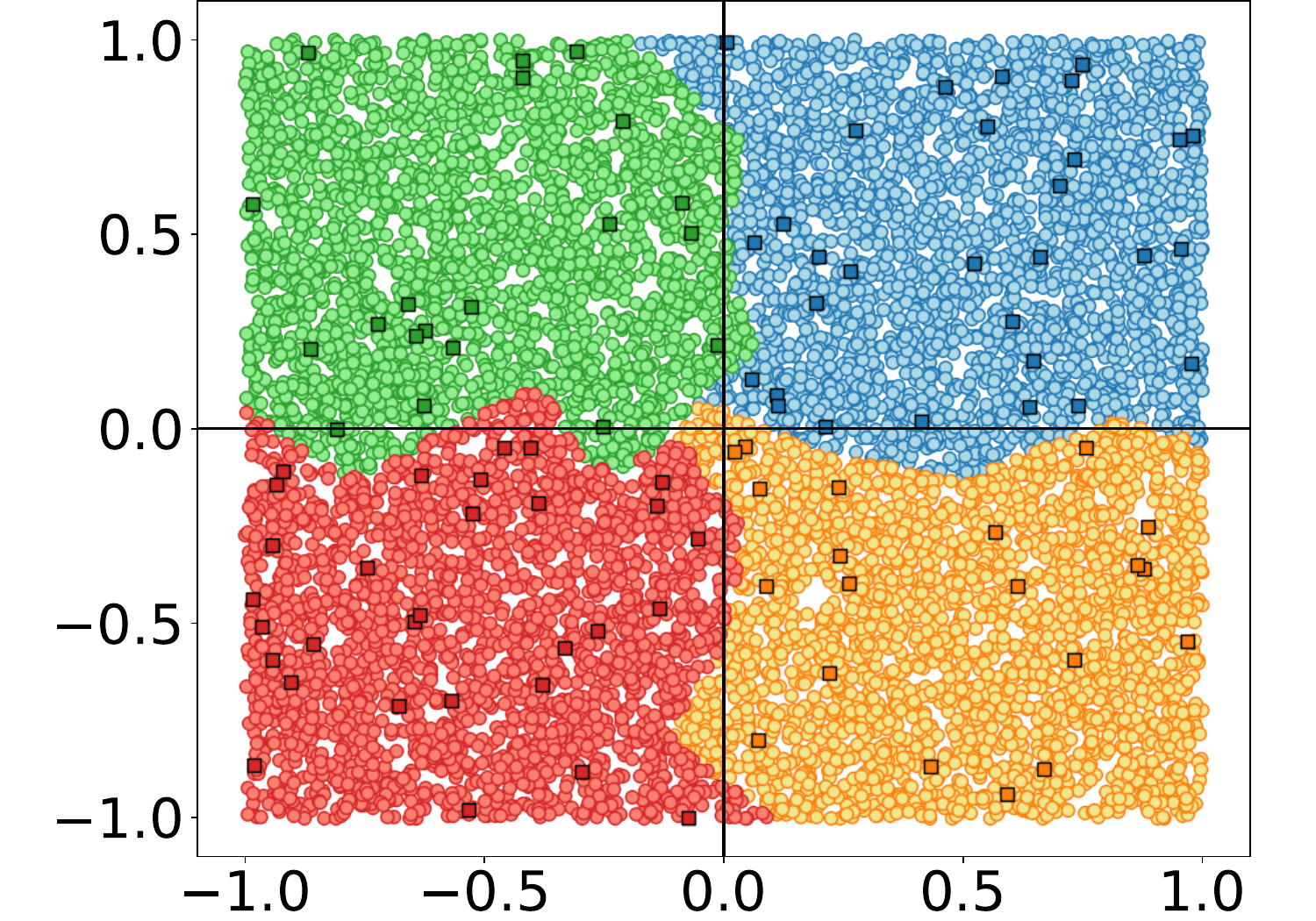}
      \end{minipage}
\end{figure}

\begin{figure}[h]
      \begin{minipage}[h]{0.32\linewidth}
         \includegraphics[keepaspectratio, width=\linewidth]{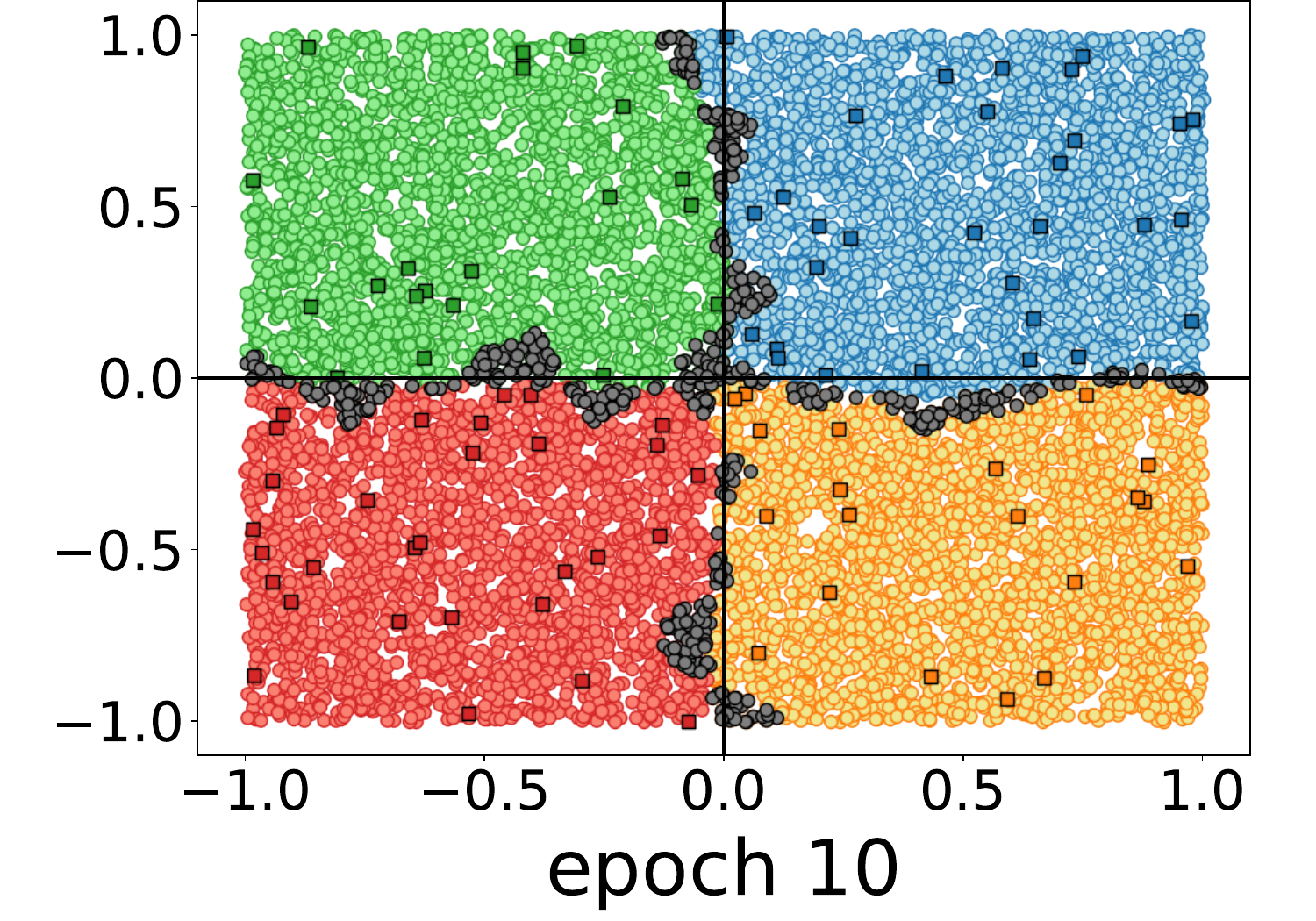}
      \end{minipage}
      \hspace{0.01\linewidth}
      \begin{minipage}[h]{0.32\linewidth}
         \includegraphics[keepaspectratio, width=\linewidth]{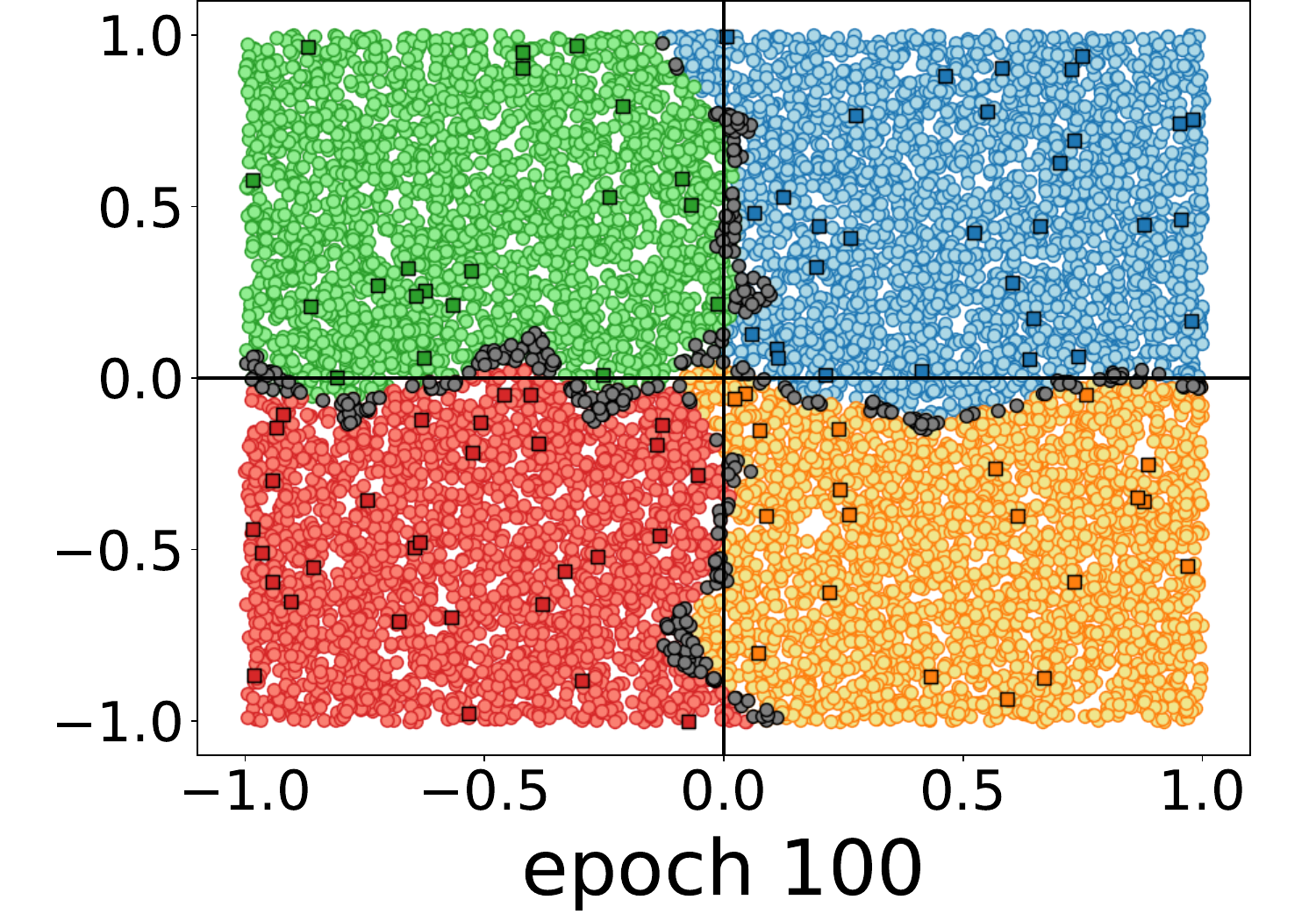}
      \end{minipage}
      \hspace{0.01\linewidth}
      \begin{minipage}[h]{0.32\linewidth}
         \includegraphics[keepaspectratio, width=\linewidth]{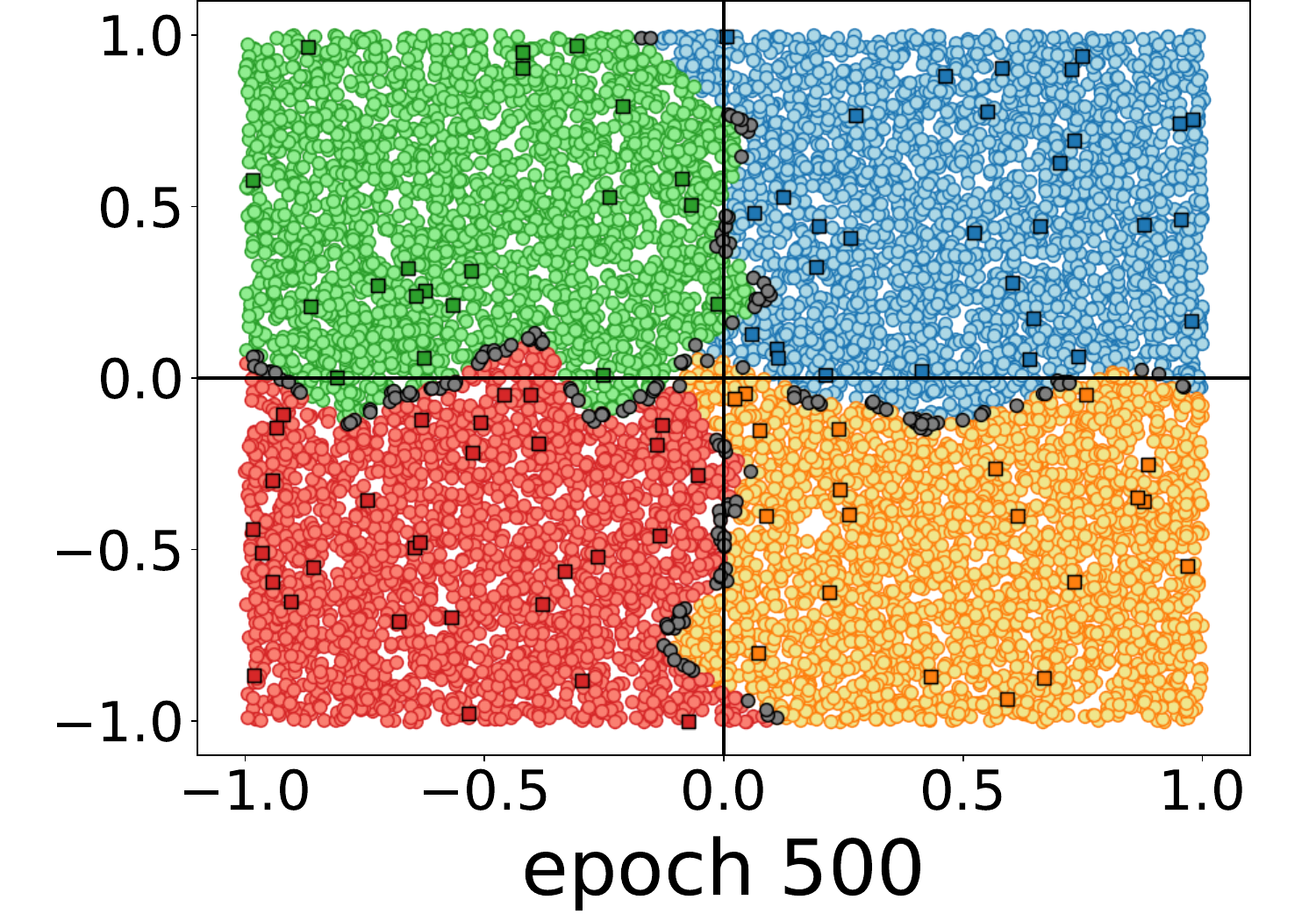}
      \end{minipage}
   \caption{Effect of our proposed method. The upper figures show the \textit{query vectors} colored by the \textit{predicted} cluster by the neural network. The bottom figures show the wrong cluster choices in gray. From left to right, border lines that the neural network predicts are fitting to the ground truth as the training progresses, and the number of wrong cluster choices decreases.}
   \label{fig_2d_training}
\end{figure}

Therefore, to improve the accuracy to choose correct cluster is the fundamental challenge. We attempt to accurately predict the complex border lines by using a neural network that is trained with the objective to choose the correct cluster.
We use simple three layer MLP. Input dimension is equal to the dimension of query and key vectors, and output dimension is equal to the number of clusters, and the dimension of hidden layer is set to 128 in this experiment.
The query vectors for training are sampled independently every epoch from the same distribution. The ground truth cluster is searched by exhaustive search for each training query.
Using those pair samples of query and ground truth cluster as training data, we train the neural network with cross-entropy loss in supervised manner. Note that this is a data-dependent method because we look into the clustered key vectors for generating the ground truth labels.
Figure~\ref{fig_2d_training} shows that prediction by the neural network approaches the correct border lines of clusters as training proceeds. 
This results indicate that we can improve the accuracy to choose correct cluster by employing a data-dependently trained neural network.

\subsection{Experiment results}
\label{experiment results}

In this section, we describe the experiment using SIFT~\citep{PQ, SIFT1B} and CLIP~\citep{CLIP} data for demonstrating that our proposed method is useful for realistic data.

\textbf{Dataset.} For SIFT1M, we use one million 128-dimensional SIFT1M base data as key vectors. Another one million data are sampled from SIFT1B base data and used as query vectors for training. SIFT1B query data are used as query vectors for test. Euclidean distance is employed as metrics. \\
For CLIP, we extracted feature vectors from 1.28 million ImageNet~\citep{ImageNet} training data with ViT B/16 model~\citep{ViT}. Although the dimension of the feature vector of the model is 512, we use the first 128 dimension for our experiment. We split it into 0.63 million, 0.64 million, and 0.01 million for key vectors, training query vectors, and test query vectors, respectively. Cosine similarity is employed as metrics. 

\textbf{Comparison with conventional methods.} We compare our method with two conventional methods. 
The first one is the exhaustive method where key vectors are partitioned by \textit{k}-means in the index building phase, and the distances of a query to the representative vectors of all the clusters partitioned by \textit{k}-means are calculated and the cluster corresponding to the closest representative vector is chosen in the search phase. The second one is SPANN~\citep{SPANN}. For SPANN, we build the index by the algorithm implemented in SPANN, which includes partitioning process. Since SPANN proposes multiple cluster assignment for improving recall, we set the \texttt{ReplicaCount} to its default value 8, which means one key vector is contained by at most 8 clusters. As described in Section~\ref{metrics_memory}, we set the number of clusters for all the methods including our proposed method to 1000 for fair comparison.

\textbf{Neural network structure.} We employ the three-layer MLP. For fair comparison, we carefully design the neural network so as not to require more memory usage than the conventional methods. As described in Section~\ref{metrics_memory}, if we employed more memory budget, we could significantly improve the recall just by increasing the number of clusters using that memory budget. Concretely, we set the size of hidden layer to 128, which is the same as the vector dimension. Then, the number of parameters of the output layer becomes the dominant among three layers and it is 128 $\times$ 1000, where 1000 is the number of clusters. The memory usage for this output layer is the same as that of the representative vectors of all the clusters which is required for the conventional methods.\\
Regarding $T_c$ for the neural network inference, the computing in the largest last layer is very close to that of distance calculations between query vectors and all the representative vectors, and the latency of this computation is much smaller than the latency for fetching vectors as shown in Table~\ref{latency}. So still the $T_b$ is dominant even employing the neural network for search.

\textbf{Training overview.} We train the neural network for 150 epochs with AdamW~\citep{ADAMW} optimizer. The batch size is set to 1000.
In order to avoid overfitting, we add some noise sampled from normal distribution every iteration to training data.
Every 50 epochs, some key vectors are added to another existing cluster. We call this process as duplication.

\textbf{Detail of duplication process.} For a training query vector, if any of top-$k_d$ clusters predicted by the neural network does not contain the ground truth key vector, the pair of top-1 cluster and the ground truth key vector is marked as a candidate pair for duplication. After checking all the training query vectors, we additionally put the key vector into the cluster for the most frequently marked $r_d$\% pairs. $k_d$ and $r_d$ are hyperparameters and set to 4 and 20 in default, respectively.

\textbf{Loss function.} We compare three loss functions. The first one is naive cross-entropy loss (CE). Although multiple clusters can contain the nearest key vector after duplication, we need to pick only one cluster as a ground truth because CE can not handle multiple positive labels. We use initial ground truth cluster as only one positive all the time. The second one is the modified version of cross-entropy loss (MCE) that picks one cluster where the neural network gives the largest score among the positive clusters. The third one is binary cross-entropy (BCE) loss that can handle multiple positive labels.

\begin{figure}[h]
   \begin{minipage}[h]{0.49\linewidth}
      \includegraphics[keepaspectratio, width=\linewidth]{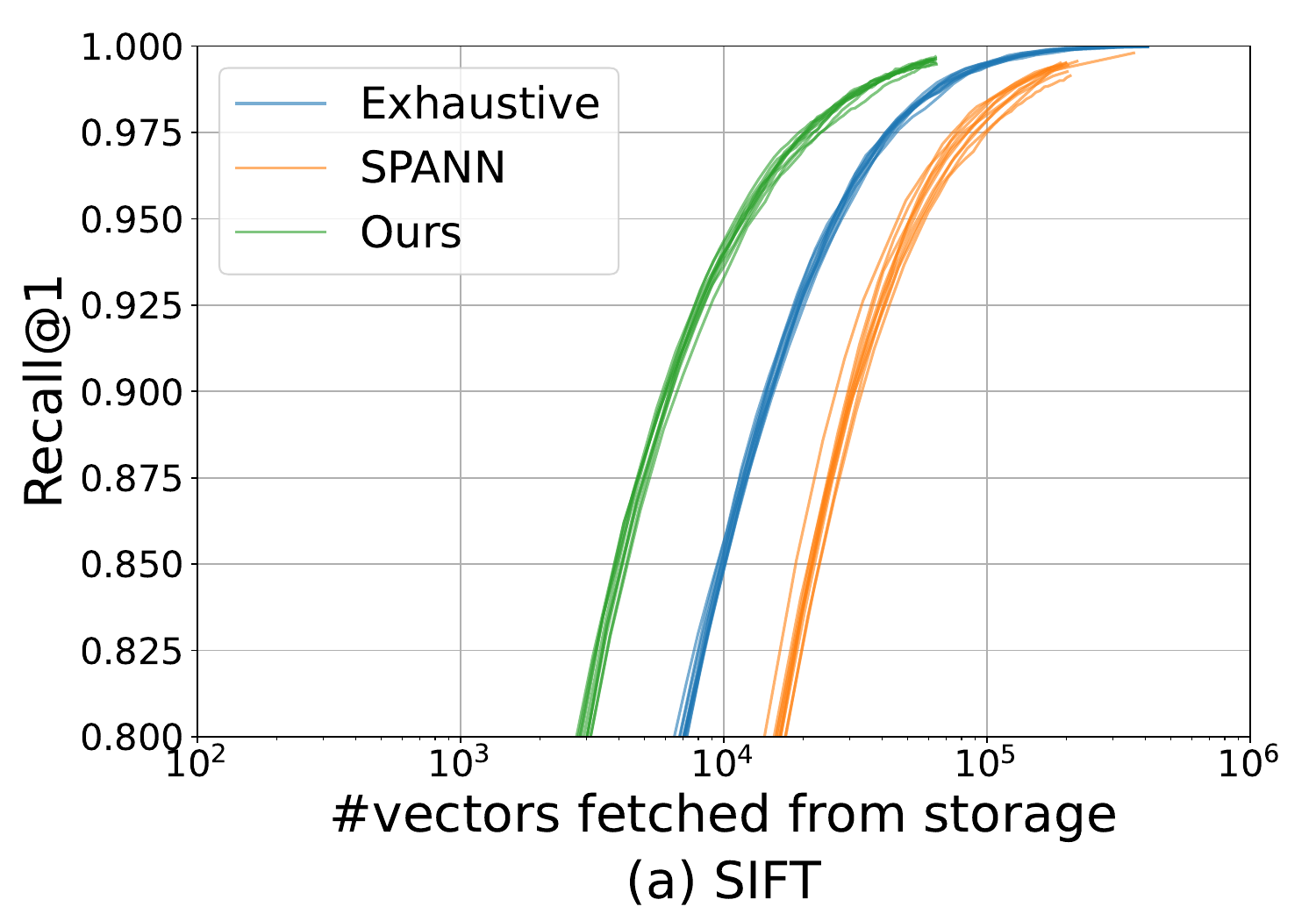}
   \end{minipage}
   \hspace{0.01\linewidth}
   \begin{minipage}[h]{0.49\linewidth}
      \includegraphics[keepaspectratio, width=\linewidth]{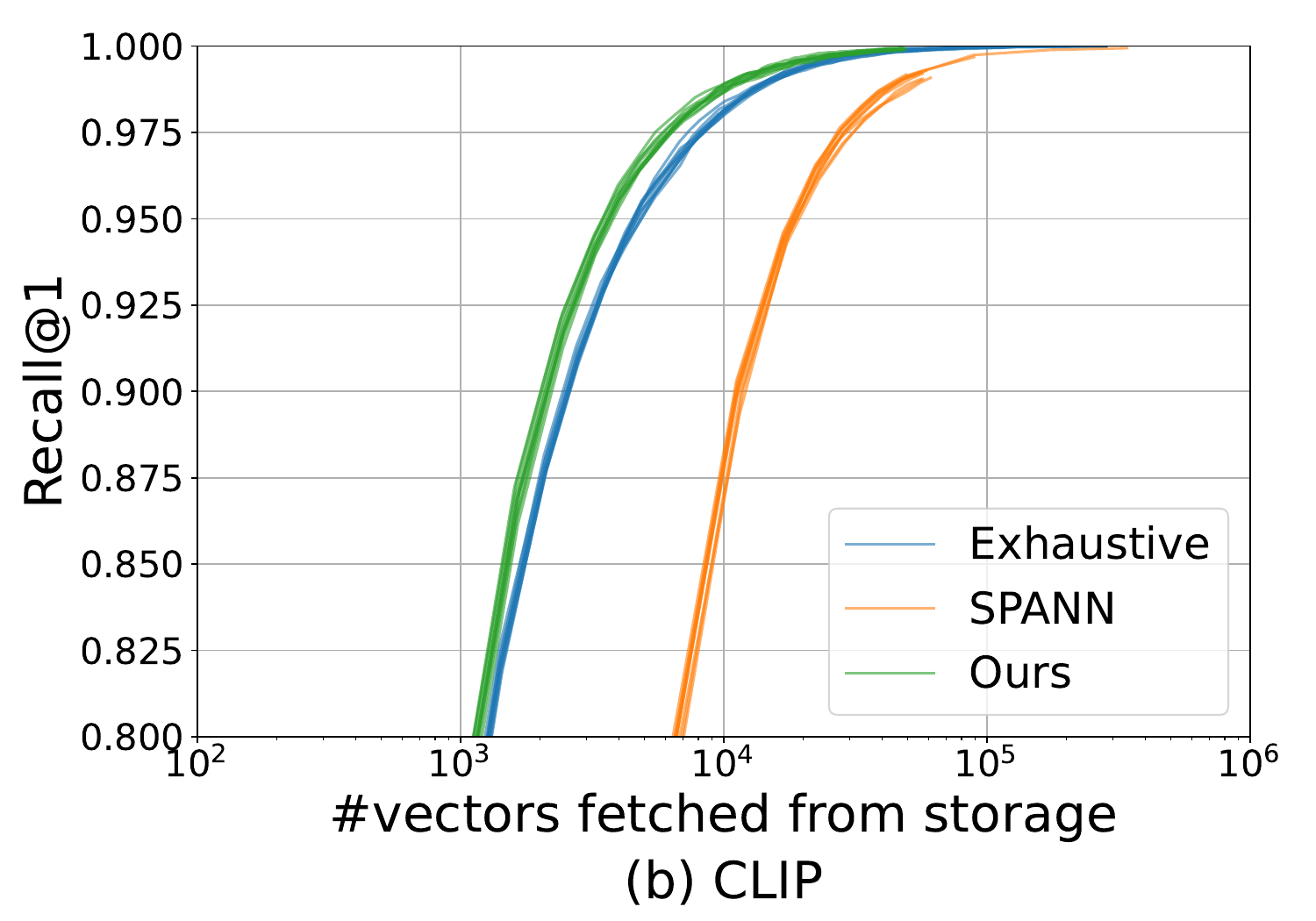}
   \end{minipage}
\caption{Recall@1 vs the number of vectors fetched from storage.}
\label{fig_experiment}
\end{figure}

\textbf{Results.} The results are shown in Figure~\ref{fig_experiment}, Table~\ref{comparison_sift}, and Table~\ref{comparison_clip}. The figure includes the results of 10 trials each. The tables show the average and the standard deviation values of the 10 trials. Our proposed method achieves the smallest number of vectors fetched from storage under any recall value, which means that our method will provide the smallest mean latency when the latency for storage access is dominant. Under 90\% recall on SIFT data, the number of vectors read from storage is 58\% and 80\% smaller than that of the exhaustive method and SPANN, respectively. Also for CLIP data, steady improvement is obtained.

\begin{table}[H]
\centering
\begin{tabularx}{103mm}{lD{.}{\pm}{5}D{.}{\pm}{5}D{.}{\pm}{5}}
   \toprule
   \textbf{Method} & \multicolumn{1}{c}{\textbf{Recall@1=90\%}} & \multicolumn{1}{c}{\textbf{Recall@1=95\%}} & \multicolumn{1}{c}{\textbf{Recall@1=99\%}} \\
   \midrule
   Exhaustive & 14900.~~392 & 26683.~~788 & 72258.~~3158 \\
   SPANN      & 30729.1749 & 52568.3691 & 151419.16055 \\
   Ours       & \textbf{6165}.~~\textbf{223} & \textbf{11932}.~~\textbf{525} & \textbf{39077}.~~\textbf{2154} \\
   \bottomrule
\end{tabularx}
\caption{Comparison of the number of fetched vectors under given recall values on SIFT data.}
\label{comparison_sift}
\end{table}

\begin{table}[H]
   \centering
   \begin{tabularx}{103mm}{lD{.}{\pm}{5}D{.}{\pm}{5}D{.}{\pm}{5}}
      \toprule
      \textbf{Method} & \multicolumn{1}{c}{\textbf{Recall@1=90\%}} & \multicolumn{1}{c}{\textbf{Recall@1=95\%}} & \multicolumn{1}{c}{\textbf{Recall@1=99\%}} \\
      \midrule
      Exhaustive & 2547.~~47 & 4655.127 & 15067.~~470 \\
      SPANN      & 11397.337 & 18531.437 & 49689.4442 \\
      Ours       & \textbf{2114}.~~\textbf{62} & \textbf{3625}.\textbf{110} & \textbf{11466}.~~\textbf{654} \\
      \bottomrule
   \end{tabularx}
   \caption{Comparison of the number of fetched vectors under given recall values on CLIP data.}
   \label{comparison_clip}
\end{table}

\subsection{Ablation study}
\label{ablation}

\subsubsection{Effect of each ingredient}

Table~\ref{tbl_ablation} shows how much each ingredient of our proposed method improves the metrics. We report the average and standard deviation values of the number of fetched vectors across 10 trials for each condition. 
(a) to (c) compare the loss functions for training. The both MCE and CE show good performance but CE is better in R@1$\le$0.95 and MCE is better in R@1$=$0.99. BCE loss deteriorates the performance and the increase in the variation is observed. (d) is the conventional exhaustive method using linear search for choosing clusters and employs neither neural networks nor duplication. (e) employs only neural network and (f) employs only duplication. By comparing (a,d,e,f), the both neural network and duplication contribute to improving performance. (g) shows the effect of the hyperparameter $k_d$ explained in Section~\ref{experiment results}. By increasing $k_d$, the number of fetched vectors decreases. In (h), we execute duplication only once after 150-epoch training is completed. The result is worse than that in the default setting where duplication is executed every 50 epochs. This indicates that executing duplication process between training can relax the complexity of the border lines of clusters and help the neural network to fit them. (i) shows the result when we use the clustering information obtained after 150 epoch training and duplication in (a) setting, but choose the cluster to be fetched by using linear search across the updated centroid vectors of all the clusters. This shows executing search with the neural network inference is advantageous. From this experiment, we can see that although even only each ingredient of our proposed method can significantly reduce the number of fetched vectors under a given recall compared to the conventional method, further improvement is obtained by combining them.

\begin{table}[h]
   \centering
   \begin{tabularx}{\linewidth}{lccccD{.}{\pm}{5}D{.}{\pm}{5}D{.}{\pm}{5}}
      \toprule
         & Training & \multicolumn{2}{c}{Duplication} & Search & \multicolumn{3}{c}{\#vectors fetched from storage} \\
         & (loss fn.) & $k_d$=4 & iter. & by NN & \multicolumn{1}{c}{R@1=0.9} & \multicolumn{1}{c}{R@1=0.95} & \multicolumn{1}{c}{R@1=0.99} \\ 
      \midrule
         a. Ours & \checkmark (CE) & \checkmark & \checkmark & \checkmark & \textbf{6165}.~~\textbf{223} & \textbf{11932}.~~\textbf{525} & 39077.2154 \\
         b. Ours & \checkmark (MCE) & \checkmark& \checkmark & \checkmark & 7724.~~588 & 13530.~~686 & \textbf{38528}.\textbf{1981} \\
         c. Ours & \checkmark (BCE) & \checkmark& \checkmark & \checkmark & 10504.1117 & 19967.1297 & 55734.3031 \\
         d. & \textbf{No train} & \multicolumn{2}{c}{\textbf{No dup.}} & \textbf{linear} & 14900.~~392 & 26683.~~788 & 72258.3158 \\
         e. & \checkmark (CE) & \multicolumn{2}{c}{\textbf{No dup.}} & \checkmark & 7305.~~210 & 13922.~~350 & 43177.1234 \\
         f. & \textbf{No train} & \checkmark & \checkmark & \textbf{linear} & 7222.~~275 & 13635.~~520 & 41969.1466 \\
         g. & \checkmark (CE) & $k_d$=1 & \checkmark & \checkmark & 6803.~~214 & 14008.~~299 & 44635.2373 \\
         h. & \checkmark (CE) & \checkmark& \textbf{-} & \checkmark & 7223.~~254 & 13956.~~718 & 40894.2445 \\
         i. & \checkmark (CE) & \checkmark & \checkmark & \textbf{linear} & 8112.~~310 & 15134.~~686 & 45768.2396 \\
      \end{tabularx}
   \caption{Effect of each ingredient.}
   \label{tbl_ablation}
\end{table}

\subsubsection{Building index by SPANN}

In our experiment, we use \textit{k}-means for partitioning, but it is an exhaustive method and can take too much time for larger dataset. 
In order to confirm that \textit{k}-means is not a necessary component of our method, we apply the algorithm in SPANN to execute partitioning.
As a result, the number of fetched vectors under 90\% recall significantly improves from 30729$\pm$1749 to 7372$\pm$162. This indicates that we may utilize a fast algorithm such as SPANN for clustering instead of exhaustive k-means when our proposed method is applied to larger dataset.

\section{Limitations and future works}

Since the discussion in this paper assumes that the condition that the mean latency for search is determined by storage access time holds true, the discussion in this paper may be invalid if this condition is not satisfied.

For CLIP data, the improvement over the exhaustive method is steady but marginal as shown in Figure~\ref{fig_experiment}. The difference of the amount of improvement between SIFT and CLIP may come from the difference of how well the training data reproduce the query distribution.
This means the effectiveness of the proposed method could be limited under the condition where the query distribution is close to uniform and not predictable. Although this is a common issue in almost all of the ANN methods, it remains future work to address such difficult use case.

Our proposed method has a couple of hyperparameters. Although we show some of their effect in Section~\ref{ablation}, thorough optimization is a future work. It may dependent on data distribution and required recall value. However, it is not difficult to find the acceptable values for hyperparameters that provide at least better performance than the exhaustive method.

Another apparent remaining future work is to apply the proposed method to larger datasets such as billion-scale or trillion-scale ones. However, we believe that foundings and direction we reveal in this paper will be also useful for them.

\section{Conclusion}
We investigated the requirement to improve the recall and latency tradeoff of large scale approximate nearest neighbor search under the condition where the key vectors are stored in storage devices with large capacity and large read latency. We pointed out that in order to achieve it, we need to reduce the number of vectors fetched from storage devices during search. Then, it is required to choose correct clusters containing the nearest neighbor key vector to a given query vector with high accuracy. We proposed to use neural networks to predict the correct cluster. By employing our proposed method, we achieved to reduce the number of vectors to read from storage by more than 58\% under 90\% recall on SIFT1M data compared to the conventional methods.


\bibliography{nn_ann}
\bibliographystyle{iclr2023_conference}


\end{document}

%% file: math_commands.tex
\usepackage{amsmath,amsfonts,bm}

\def\eqref#1{equation~\ref{#1}}

\def\1{\bm{1}}

\DeclareMathAlphabet{\mathsfit}{\encodingdefault}{\sfdefault}{m}{sl}
\SetMathAlphabet{\mathsfit}{bold}{\encodingdefault}{\sfdefault}{bx}{n}

